\documentclass[10pt,twocolumn,letterpaper]{article}

\usepackage{cvpr}
\usepackage{times}
\usepackage{epsfig}
\usepackage{graphicx}
\usepackage{amsmath}
\usepackage{amssymb}
\usepackage{caption}
\usepackage{url}
\usepackage[breaklinks=true,bookmarks=false]{hyperref}

\cvprfinalcopy 

\def\cvprPaperID{108} 

\ifcvprfinal\fi

\begin{document}

\title{WiCV 2021: The Eighth Women In Computer Vision Workshop}

\author{
 Arushi Goel$^1$, Niveditha Kalavakonda$^2$, Nour Karessli$^3$, Tejaswi Kasarla$^4$, \\Kathryn Leonard$^5$, Boyi Li$^6$, Nermin Samet$^7$, Ghada Zamzmi$^8$ \\\\ $^1$University of Edinburgh, $^2$University of Washington, $^3$Zalando SE,\\$^4$Bosch Research, $^5$Occidental College,  $^6$Cornell University, \\ $^7$Middle East Technical University, $^8$National Institutes of Health\\
 \tt\small wicvcvpr2021-organizers@googlegroups.com
}

\maketitle

\begin{abstract}
\thispagestyle{empty}
In this paper, we present the details of Women in Computer Vision Workshop - WiCV 2021, organized alongside the virtual CVPR 2021.  It provides a voice to a minority (female) group in the computer vision community and focuses on increasing the visibility of these researchers, both in academia and industry. WiCV believes that such an event can play an important role in lowering the gender imbalance in the field of computer vision. WiCV is organized each year where it provides a)~opportunity for collaboration between researchers from minority groups, b)~mentorship to female junior researchers,  c)~financial support to presenters to overcome monetary burden and d)~large and diverse choice of role models, who can serve as examples to younger researchers at the beginning of their careers. In this paper, we present a report on the workshop program, trends over the past years, a summary of statistics regarding presenters, attendees, and sponsorship for the WiCV 2021 workshop.

\end{abstract}

\section{Introduction}
While excellent progress has been made in a wide variety of computer vision research areas in recent years, similar progress has not been made in the increase of diversity in the field and the inclusion of all members of the computer vision community. Despite the rapid expansion of our field, females still only account for a small percentage of the researchers in both academia and industry. Due to this, many female computer vision researchers can feel isolated in workspaces which remain unbalanced due to the lack of inclusion.

The Women in Computer Vision workshop is a gathering for both women and men working in computer vision. It aims to appeal to researchers at all levels, including established researchers in both industry and academia (e.g. faculty or postdocs), graduate students pursuing a Masters or PhD, as well as undergraduates interested in research.  This aims to raise the profile and visibility of female computer vision researchers at each of these levels, seeking to reach women from diverse backgrounds at universities and industry located all over the world.

There are three key objectives of the WiCV workshop.
The first to increase the WiCV network and promote interactions between members of this network, so that female students may learn from professionals who are able to share career advice and past experiences. A mentoring banquet is run alongside the workshop. This provides a casual environment where both junior and senior women in computer vision can meet, exchange ideas and even form mentoring or research relationships.

The workshop's second objective is to raise the visibility of women in computer vision. This is done at both the junior and senior levels. Several senior researchers are invited to give high quality keynote talks on their research, while junior researchers are invited to submit recently published or ongoing works with many of these being selected for oral or poster presentation through a peer review process. This allows junior female researchers to gain experience presenting their work in a professional yet supportive setting. We strive for diversity in both research topics and presenters' backgrounds. The workshop also includes a panel, where the topics of inclusion and diversity can be discussed between female and male colleagues.

Finally, the third objective is to offer junior female researchers the opportunity to attend a major computer vision conference which they otherwise may not have the means to attend. This is done through travel grants awarded to junior researchers who present their work in the workshop via a poster session. These travel grants allow the presenters to not only attend the WiCV workshop, but also the rest of the CVPR conference. 

\section{Workshop Program}
\label{program}

The workshop program consisted of 4 keynotes, 4 oral presentations, 36 poster presentations, a panel discussion, and a mentoring session. As with previous years, our keynote speakers were selected to have diversity among topic, background, whether they work in academia or industry, as well as their seniority. It is crucial to provide a diverse set of speakers so that junior researchers have many different potential role models who they can relate to in order to help them envision their own career paths.

The workshop schedule was as follows:
\begin{itemize}
\item Introduction
\item Invited Talk 1: Zeynep Akata (University of Tübingen), \textit{Explainability and Compositionality for Visual Recognition}
\item Invited Talk 2: Diane Larlus (Naver Labs Europe), \textit{Lifelong Visual Representation Learning}
\item Oral Session
\begin{itemize}
\item Naina Dhingra (ETH Zurich), \textit{BGT-Net: Bidirectional GRU Transformer Network for Scene Graph Generation}
\item Rita Pucci (Univeristy of Udine), \textit{Collaborative Image and Object Level Features for Image Colourisation}
\item Tahani Madmad (Université Catholique de Louvain), \textit{CNN-based morphological decomposition of X-ray images for defects and local structures contrast enhancement}
\item Purvi Goel (Facebook), \textit{On the Robustness of Monte Carlo Dropout Trained with Noisy Labels}
\end{itemize}

\item Poster Session (repeated 12 hours later)
\item Invited Talk 3: Natalia Neverova (Facebook AI Research), \textit{Towards universal predictors for object correspondences}
\item Invited Talk 4: Olga Russakovsky (Princeton University), \textit{Perception, interaction and fairness: key components of visual recognition}
\item Panel Discussion
\item Closing Remarks
\item Mentoring Session (repeated 12 hours later)
\begin{itemize}
\item Speaker: Fatma Guney (Koç University)
\item Speaker: Aishwarya Agrawal (MILA, University of Montreal)
\end{itemize}
\end{itemize}

\subsection{Virtual Setting}
This year, the organization has been slightly modified as CVPR 2021 was held remotely. We made sure to make the virtual WiCV workshop as engaging and interactive as possible. We used OhYay as our virtual platform for the main workshop and zoom for the mentoring dinner, which provided us the opportunity to conduct talks and panel discussion seamlessly. Each poster session was allocated a private room, allowing easy display of a poster or video and live discussion amongst presenters and attendees. The workshop was set up to have two poster presentation sessions with a 12 hours difference between them to cope with different time zones. This maximised the amount of people able to attend the workshop and see the accepted works. All the presented keynotes and talks have been uploaded to our website to allow asynchronous viewing after the workshop was over.

\section{Workshop Statistics}

Originally, the first workshop for WiCV was held in conjunction with CVPR 2015. Since then, the participation rate and number quality of submissions to WiCV have been steadily increasing. 
Following the examples from the editions held in previous years \cite{Akata18,Amerini19,Demir18,doughty2021wicv}, we were encouraged to collect the top quality submissions into workshop proceedings. By providing oral and poster presenters with the opportunity to publish their work in the conference's proceedings, we believe that the visibility of female researchers will be further increased.
This year, the workshop was held as a half day virtual gathering over OhYay and Zoom. Senior and junior researchers were invited to present their work, and poster presentations are included as already described in the previous Section \ref{program}.\\

The organizers for this year WiCV workshop are working in both academia and industry from various institutions located in six different time zones, with up to twelve and half hours difference. Their miscellaneous backgrounds and research areas have pledged the organizing committee a diverse perspective. Their research interests in computer vision and machine learning include video understanding, computational geometry, representation learning, uncertainty modelling, optimization, multi-modal and semi-supervised learning in different industrial application areas such as healthcare, and fashion.

\begin{figure}[h]
\centering
\includegraphics[width=1\linewidth]{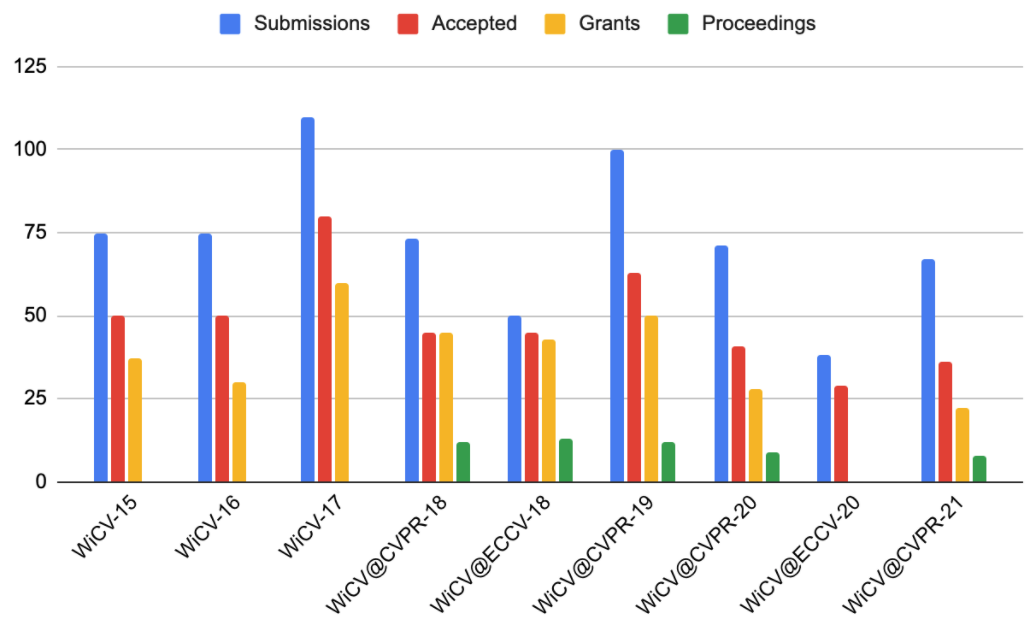}
\captionof{figure}{\textbf{WiCV Submissions.} The number of submissions over the past years of WiCV.}
\label{fig:sub}
\end{figure}

This year we had 67 high quality submissions from a wide range of topics and institutions. This is slightly reduced from previous year due to the global pandemic. It is on par WiCV@CVPR18, the previous CVPR edition which also had an ECCV\cite{Akata18} edition in the same year. The most popular topics were object recognition and deep learning and convolutional neural networks followed by video understanding, images and language and vision for robotics. 
Over all submissions, around 6\% were selected to be presented as oral talks and 54\% were selected to be presented as posters. Within the accepted submissions, 4 papers were selected to be included in the workshop's proceedings. The comparison with previous years is presented in Figure~\ref{fig:sub}. With the great effort of an interdisciplinary program committee consisting of 34 reviewers, the submitted papers were evaluated and received valuable feedback.

Although this edition was exceptionally held remotely, we have persisted WiCV tradition of last year's workshops \cite{Akata18,Amerini19,Demir18,doughty2021wicv} in providing grants to help the authors of accepted submissions participate in the workshop. The grants covered the conference workshop registration fees and daycare sponsorships (if any) for all the authors of accepted submissions who requested funding. 

\begin{figure}
\centering
\includegraphics[width=1\linewidth]{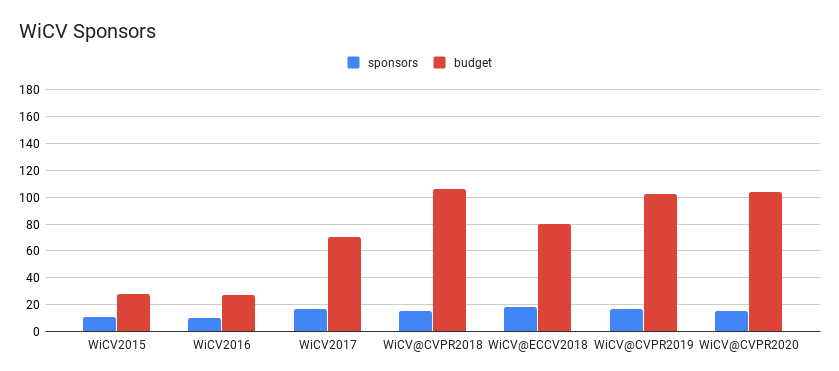}
\captionof{figure}{\textbf{WiCV Sponsors.} The number of sponsors and the amount of sponsorships for WiCV. The amount is expressed in US dollar (USD).}
\label{fig:spo}
\end{figure}

The total amount of sponsorship this year is \$70,000 USD with 14 sponsors, reaching a very good target. In Figure~\ref{fig:spo} you can find the details with respect to the past years. The majority of this sponsorship was spent on registration stipends and other daycare funding needs of participants. 

\section{Conclusions}
WiCV at CVPR 2021 has continued to be a valuable opportunity for presenters, participants and organizers in providing a platform to bring the community together. It continues to overcome the existing issue of gender balance prevailing around us and we hope that it has played an important part in making the community even stronger. Even as a virtual workshop, it provided an opportunity for people to connect from all over the world from their personal comforts. With distance and commute no longer an issue, virtual workshops brought the community even closer. With a high number of paper submissions and even higher number of attendees, we foresee that the workshop will continue the marked path of previous years and foster stronger community building with increased visibility, providing support, and encouragement inclusively for all the female researchers in academia and in industry.

\section{Acknowledgments}
First of all, we would like to thank our sponsors. We are very grateful to our other Platinum sponsors: Toyota Research Institute, Microsoft, Apple and Google Research. We would also like to thank our Gold sponsor: Pinterest, Zalando, Snapchat and Amazon; Silver Sponsors: Facebook, Intel AI and DeepMind; Bronze sponsors: Meshcapade, Argo AI and Disney Research. We would also like to thank Occidental College as our fiscal sponsor, which donated employee knowledge and time to process our sponsorships and travel awards. We would also like to thank and acknowledge the organizers of WiCV at CVPR 2020, without the information flow and support from the previous WiCV organizers, this WiCV would not have been possible. We would also like to acknowledge CVPR 2020 Workshop Chairs Hazel Doughty, Srishti Yadav, Arsha Nagrani and Azadeh Monasher for answering all questions concerns timely. Huge thanks also go to the CVPR committee who made a mammoth effort to make a virtual CVPR possible. Finally, we would like to acknowledge the time and efforts of our program committee, authors, reviewers, submitters, and our prospective participants for being part of WiCV network community.

\section{Contact}
\noindent \textbf{Website}: \url{https://sites.google.com/view/wicvcvpr2021/home}\\
\textbf{E-mail}: wicvcvpr2021-organizers@googlegroups.com\\
\textbf{Facebook}: \url{https://www.facebook.com/WomenInComputerVision/}\\
\textbf{Twitter}: \url{https://twitter.com/wicvworkshop}\\
\textbf{Google group}: women-in-computer-vision@googlegroups.com \\

{\small
\bibliographystyle{ieee}
\bibliography{egbib}
}

\end{document}